\documentclass{article} 
\PassOptionsToPackage{numbers,sort&compress}{natbib}
\usepackage{iclr2027_conference,times}
\setcitestyle{numbers,square,citesep={,},aysep={,}}

\usepackage{amsmath,amsfonts,bm}

\def\eqref#1{equation~\ref{#1}}

\def\1{\bm{1}}

\DeclareMathAlphabet{\mathsfit}{\encodingdefault}{\sfdefault}{m}{sl}
\SetMathAlphabet{\mathsfit}{bold}{\encodingdefault}{\sfdefault}{bx}{n}

\usepackage[utf8]{inputenc}
\usepackage[T1]{fontenc}
\usepackage{hyperref}
\usepackage{url}
\usepackage{booktabs}
\usepackage{amsfonts}
\usepackage{amssymb}
\usepackage{amsmath}
\usepackage{amsthm}
\usepackage{mathtools}
\usepackage{microtype}
\usepackage{xcolor}
\usepackage{graphicx}
\usepackage{subcaption}
\usepackage{enumitem}

\usepackage{amsmath,amsthm,amssymb}
\usepackage{algorithm,algorithmic}
\allowdisplaybreaks

\newcommand{\diag}{\mathrm{diag}}

\newcommand{\defeq}{:=}

\newcommand{\vQ}{\mathbf{Q}}
\newcommand{\vK}{\mathbf{K}}
\newcommand{\vV}{\mathbf{V}}

\newcommand{\vdQ}{\mathbf{dQ}}
\newcommand{\vdK}{\mathbf{dK}}
\newcommand{\vdV}{\mathbf{dV}}

\newcommand{\vS}{\mathbf{S}}
\newcommand{\vdS}{\mathbf{dS}}

\newcommand{\vP}{\mathbf{P}}
\newcommand{\vdP}{\mathbf{dP}}

\newcommand{\vO}{\mathbf{O}}
\newcommand{\vdO}{\mathbf{dO}}

\newcommand{\vm}{\mathbf{m}}
\newcommand{\vell}{\mathbf{\ell}}

\newcommand{\fullName}{CoWindow Attention}
\newcommand{\shortName}{CoWA}

\title{\fullName{}:\\ Full Causal Coverage Is a Collective Property}

\author{
  \parbox{\textwidth}{
    Jingze Shi$^{12}$\thanks{Equal contribution. $^1$The Hong Kong University of Science and Technology (Guangzhou), Guangzhou, China. $^2$Beijing Academy of Artificial Intelligence, Beijing, China. $^3$Université Paris Cité, Paris, France. Corresponding author: Guang Liu <liuguang@baai.ac.cn>, Yuyu Luo <yuyuluo@hkust-gz.edu.cn>.} \ \ Zhangyang Peng$^{1}$ \ \ Xianduo Li$^{2}$ \ \ Yanlin Qi$^{23}$ \ \ Xiaotian Lin$^{1}$ \ \ Haoxian Chen$^{12}$  \\[2pt]
    Liangdong Wang$^{2}$ \ \ Guang Liu$^{2}$ \ \ Yuyu Luo$^{1}$
  }
}

\iclrfinalcopy 
\begin{document}

\maketitle
\fancyhead{}
\lhead{Full Causal Coverage Is a Collective Property}

\begin{abstract}
Long-context full attention (FullAttn) repeatedly exposes the complete causal history to every attention head, creating substantial redundant computation and memory traffic even with IO-efficient dense kernels. We introduce \textbf{\fullName{} (\shortName{})}, a structured attention architecture that distributes access to the causal history across KV heads. All heads share near-diagonal and prefix-sink windows, while complementary long-range windows partition the remaining history. Their union provides full causal coverage although each head attends sparsely to distant tokens. This position-defined attention pattern requires no learned router or indexer, is used consistently during training and inference, and aligns with KV-head tensor parallelism. A window-matched ablation at 8K isolates the effect of complementary long-range allocation: \shortName{} with 100\% collective coverage reaches 89.73\% accuracy, compared with 89.97\% for FullAttn, while duplicated long-range windows perform substantially worse. Across a broader controlled associative-recall comparison with matched token budgets, \shortName{} closely tracks FullAttn as the context grows, whereas other sparse patterns lose a substantial fraction of the associations. In an attention-operator benchmark at 128K tokens on 8 GPUs with tensor parallelism, \shortName{} reduces forward and backward latency during training by $7.4\times$ and $8.6\times$ and decoding latency during inference by $3.0\times$ over FullAttn. Its per-rank peak operator memory matches FullAttn during training and is $7.6\times$ lower during decoding. Across scaling-law training from 0.6B to 14B parameters on 128 GPUs, \shortName{} closely tracks FullAttn in perplexity while reducing total training FLOPs, with a 28.5\% reduction at 14B during 32K-context training. The resulting 14B models and 32B models from separate continued training achieve comparable knowledge, reasoning, and long-context retrieval scores to FullAttn. These results show that full causal coverage can be a collective property of the head ensemble rather than a duplicated property of every head.
Our code is open-sourced at \href{https://github.com/HKUSTDial/flash-sparse-attention}{\nolinkurl{flash-sparse-attention}}.
\end{abstract}

\section{Introduction}
\label{sec:introduction}

Context lengths are expanding from thousands to hundreds of thousands of tokens~\citep{snell2024tts} to support long-document understanding~\citep{park2023generative,gemini2025}, multi-turn reasoning~\citep{hf2025openr1,deepseekai2025deepseekr1incentivizingreasoningcapability,qwen32025}, and repository-level code generation~\citep{zhang2024codeagent}.
Attention over these long contexts incurs substantial computation and memory traffic.
FlashAttention~\citep{dao2022flashattention,shah2024flashattention3} improves the IO efficiency of self-attention~\citep{vaswani2017attention} through tiling, fusion, and online softmax~\citep{milakov2018onlinesoftmax}.
However, these IO improvements leave the full attention (FullAttn) pattern unchanged: every head still attends to the full causal prefix.

Retaining direct access to this history within an attention layer does not require every head to access every historical position~\citep{zhao2025makingheadcountsparse}.
For each query, such access can be provided collectively, with every causal token position visible to at least one head.
Attention heads develop heterogeneous roles and exhibit distinct long-range and local behaviors~\citep{guo2024activedormantah,gu2024whenas,barbero2025whydl,xiao2024duoattentionel,sandovalsegura2025identifyingae,fu2026attentionsinkforgesnative}.
These observations motivate keeping recent context visible to all heads while dividing distant context among heads with complementary receptive fields.

Coordinating this access across KV heads also requires an execution structure that remains efficient during training and inference.
Sliding-window attention is simple and efficient, but applies the \emph{same finite horizon} to each head and removes direct access to more distant tokens~\citep{fu2025slidingwindowattentiontraining}.
Global tokens and fixed block patterns retain only predetermined long-range paths~\citep{child2019generating,zaheer2020big}, while dynamic approaches select tokens or blocks through content-dependent scores or routing~\citep{tang2024quest,lai2025flexprefill,li2024snapkv,zhang2023h2o,xiao2024infllm,qi2026pariskv,zhao2025infllmv2,yuan2025nativesparseattentionhardwarealigned,lu2025moba,gao2024seerattention}.
These dynamic mechanisms offer greater flexibility, but introduce a separate selection problem and often depart from the regular window structure that makes sparse attention inexpensive.
This raises the question: can distributing access to the causal history across KV heads preserve model quality and long-range retrieval while reducing the cost of training and inference?

\begin{figure}[!t]
    \centering
    \includegraphics[height=0.26\textheight]{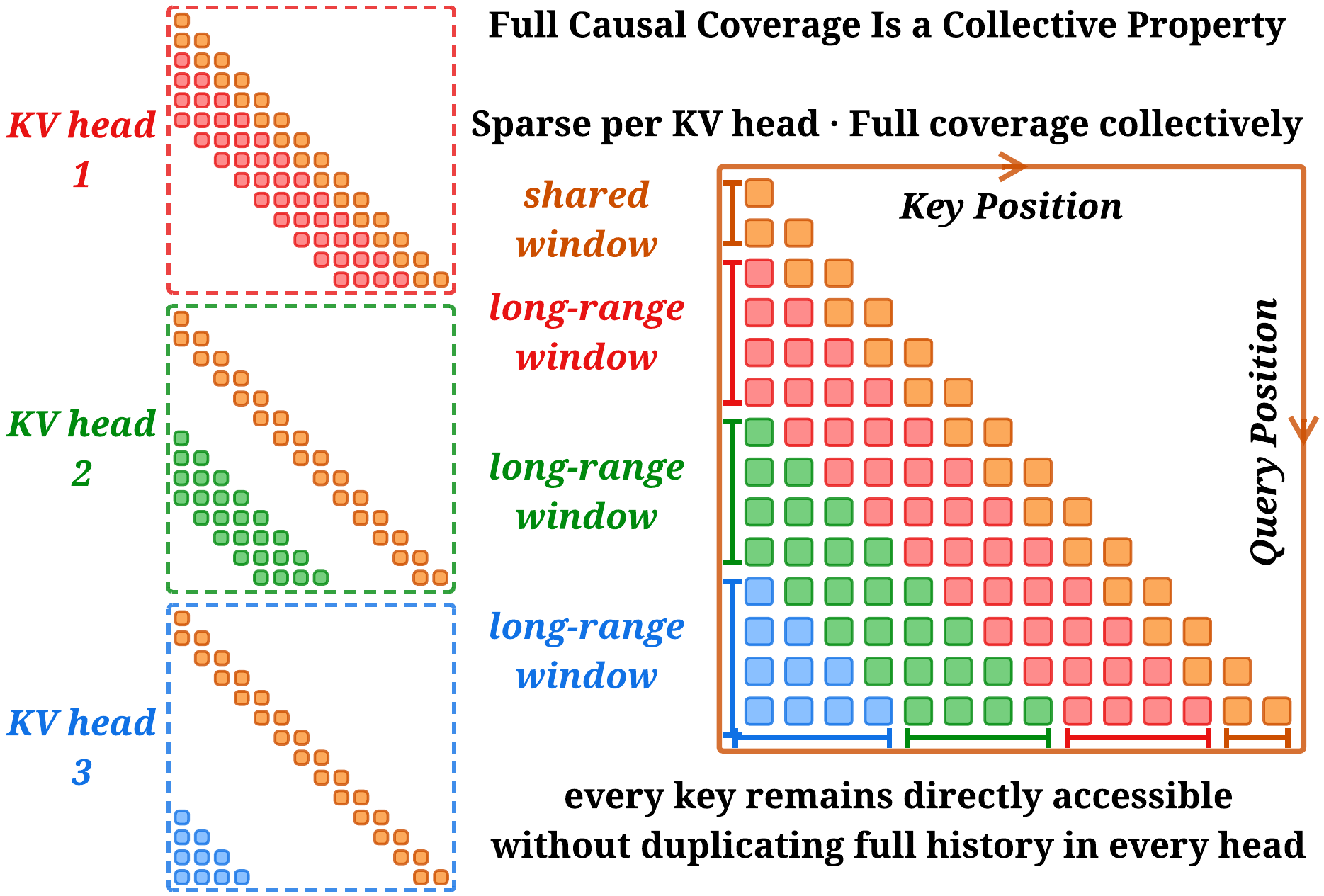}%
    \hspace{-0.095em}%
    \includegraphics[height=0.26\textheight]{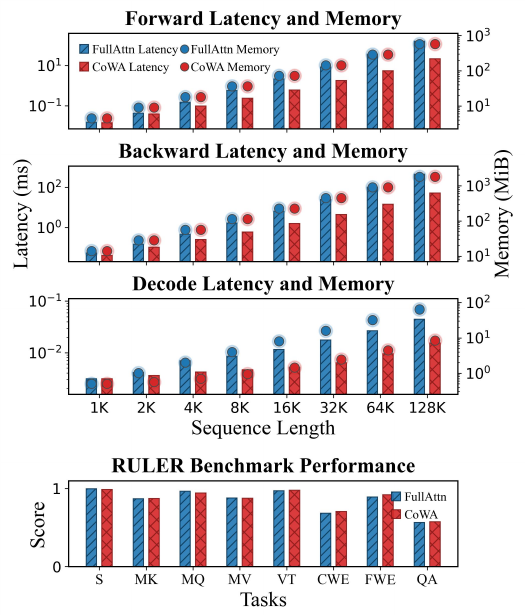}%
    \caption{
        \textbf{\fullName{} Overview.}
        \textbf{Left:} All KV heads share the near window and receive complementary long-range windows whose union covers the full causal attention region, preserving direct access to every preceding key position without duplicating the full history in every head. The shared sink window is omitted for visual clarity.
        \textbf{Right:} As sequence length grows, \shortName{} reduces forward and backward latency during training and decoding latency during inference, while matching FullAttn's peak operator memory during training and reducing it during decoding. Separate model-level evaluations show comparable performance on the RULER benchmark.
    }
    \label{fig:teaser}
\end{figure}

We introduce \textbf{\fullName{} (\shortName{})}, which distributes access to the causal history across KV heads, as illustrated in Figure~\ref{fig:teaser}.
All KV heads share a near-diagonal window that preserves direct access to recent context and a prefix-sink window that keeps the beginning of the sequence visible.
\shortName{} then partitions the remaining causal distances into complementary long-range windows and assigns one window to each KV head.
Consequently, each head remains locally dense but becomes sparse over the distant history, while together the heads can attend to every position in the causal prefix.
We call this property \emph{collective coverage}.

\shortName{} preserves full causal coverage within each attention layer while reducing duplicated long-range query-key connections across heads.
We train the input projections and output projection end to end under this sparse, position-defined attention pattern, allowing the model to learn how to combine information accessed by different heads.

The same decomposition also yields a regular execution structure.
Each head is described by a small set of contiguous windows, allowing excluded attention blocks to be omitted directly rather than discovered through a dense mask or a separate router.
A single position-defined window-construction rule governs the training forward and backward passes, inference prefill and autoregressive decoding.
Global KV-head indexing preserves complementary window assignments across tensor-parallel ranks.

We evaluate \shortName{} around two questions: whether collective coverage preserves language-model quality and long-range retrieval when no individual head observes the full history, and whether the resulting reduction in duplicated long-range access translates into practical training and inference efficiency under tensor parallelism~\citep{megatron-lm}.
A window-matched collective-coverage ablation at 8K first isolates the effect of complementary long-range allocation: \shortName{} reaches 89.73\% accuracy, compared with 89.97\% for FullAttn, while duplicated long-range windows perform substantially worse under matched per-head window widths.
Across a broader controlled associative-recall comparison with matched token budgets among sparse methods, \shortName{} closely tracks FullAttn as the context grows, whereas the other sparse patterns degrade substantially.
In the attention-operator benchmark at 128K sequence length on 8 H100 GPUs with tensor parallelism, \shortName{} reduces forward and backward latency during training by $7.4\times$ and $8.6\times$, and decoding latency during inference by $3.0\times$ relative to FullAttn.
Across scaling-law training from 0.6B to 14B parameters, \shortName{} closely tracks FullAttn in perplexity; at 14B, it reduces total training FLOPs by 3.1\% during 4K pre-training and 28.5\% during 32K long-context training.
At the model level, the resulting 14B models and separately continued-trained 32B models retain aggregate knowledge, reasoning, and long-context retrieval performance comparable to FullAttn.
Our contributions are as follows:
\begin{itemize}[leftmargin=1em]
    \item We formulate \textbf{\fullName{}}, which distributes access to the causal history through shared near and sink windows and complementary long-range windows.
    This construction provides full causal coverage while reducing duplicated long-range access across heads.
    \item We implement \shortName{} for training and inference using a position-defined window rule aligned with KV-head tensor parallelism.
    \item We evaluate the modeling and efficiency consequences of collective coverage across long-context language modeling, retrieval, kernel execution, and distributed deployment, and isolate its contribution through a window-matched ablation.
\end{itemize}

\section{Methodology}
\label{sec:methodology}

We now formalize the head-wise attention pattern underlying \shortName{} and describe its realization for training and inference under tensor parallelism.

\subsection{\fullName{} Pattern}
\label{sec:fwa_support}

The default \shortName{} pattern assigns each KV head three windows: shared near-diagonal and prefix-sink windows, and a head-specific long-range window. The long-range windows cover complementary causal-distance intervals across KV heads.

Let $N_q$ and $N_k$ denote the query and key sequence lengths, with zero-based token indices $0\le i_q<N_q$ and $0\le i_k<N_k$ throughout.
To use one definition throughout training and inference, including variable-length prefill and autoregressive decoding, we define the bottom-right-aligned causal distance
\begin{equation}
    \label{eq:fwa_distance}
    \delta(i_q,i_k)
    \defeq
    i_q+(N_k-N_q)-i_k
\end{equation}
A key is causal for query position $i_q$ exactly when $\delta(i_q,i_k)\ge 0$.
When $N_q=N_k$, this reduces to the usual distance $i_q-i_k$.

Let $H_q$ and $H_k$ be the numbers of query/output (QO) and key/value (KV) heads, with $H_q$ an integer multiple of $H_k$.
Under grouped-query attention~\citep{ainslie2023gqa}, QO head $h$ uses the key and value tensors and the window assignment of KV head $h_k=\lfloor hH_k/H_q\rfloor$.
The associated $H_q/H_k$ QO heads form a QO-head group that shares keys, values, and the visible-key set while computing attention weights from distinct queries.
We first describe the default pattern, in which all KV heads share a prefix-sink window of width $w_{\mathrm{sink}}$ and a near-diagonal window of width $w_{\mathrm{near}}$.
The remaining long-range span is divided evenly using breakpoints $b_{h_k}$:
\begin{equation}
    \label{eq:fwa_long_range_span}
    L\defeq\max(N_k-w_{\mathrm{sink}}-w_{\mathrm{near}},0)
    \qquad
    b_{h_k}\defeq
    \left\lfloor h_kL/H_k \right\rfloor,
    \quad h_k=0,\ldots,H_k
\end{equation}
For a fixed query $i_q$, write $\delta=\delta(i_q,i_k)$.
The three windows assigned to KV head $h_k$ are
\begin{equation*}
    \begin{aligned}
    \mathcal{W}^{\mathrm{near}}_{h_k}(i_q)
    &\defeq \{i_k : 0 \le \delta < w_{\mathrm{near}}\}
    && \text{(near-diagonal)}\\
    \mathcal{W}^{\mathrm{sink}}_{h_k}(i_q)
    &\defeq \{i_k : 0 \le i_k < w_{\mathrm{sink}}, \delta\ge 0\}
    && \text{(prefix-sink)}\\
    \mathcal{W}^{\mathrm{long}}_{h_k}(i_q)
    &\defeq \{i_k : w_{\mathrm{near}}+b_{h_k}\le\delta<w_{\mathrm{near}}+b_{h_k+1}\}
    && \text{(long-range)}
    \end{aligned}
\end{equation*}
Their union gives the visible-key set:
\begin{equation}
    \label{eq:fwa_support}
    \mathcal{A}_{h_k}(i_q)
    \defeq
    \mathcal{W}^{\mathrm{near}}_{h_k}(i_q)
    \cup
    \mathcal{W}^{\mathrm{sink}}_{h_k}(i_q)
    \cup
    \mathcal{W}^{\mathrm{long}}_{h_k}(i_q)
\end{equation}
Each head therefore sees the shared recent context and prefix sinks together with one long-range interval.

Algorithms~\ref{alg:fwa_forward} and~\ref{alg:fwa_backward} represent each head's windows by a four-tuple of sink width, near width, gap after the near window, and long-range width.
For the default pattern, this tuple is $w_{h_k}=(w_{\mathrm{sink}},w_{\mathrm{near}},b_{h_k},b_{h_k+1}-b_{h_k})$.
The same interface permits head-specific window widths and gaps.
Other breakpoint schedules redistribute long-range work under the attention pattern in Equation~\ref{eq:fwa_support}.
For load-balanced decoding, we use the default equal-width allocation; equal-area allocation for training is outside the scope of this work.

\begin{algorithm}[!t]
    \small
    \caption{\fullName{} Forward}
    \label{alg:fwa_forward}
    \vspace{-0.25em}
    \begin{algorithmic}[1]
        \REQUIRE Matrices $\vQ, \vO \in \mathbb{R}^{H_q \times N_q \times d_h}$ and $\vK, \vV \in \mathbb{R}^{H_k \times N_k \times d_h}$. Vector $\vell \in \mathbb{R}^{H_q \times N_q}$. Windows $w$.
        \STATE Divide $\vQ, \vO$ into QO blocks of size $B_q$ and $\vK, \vV$ into KV blocks of size $B_k$.
        \FOR{each QO head $h \in [0, \ldots, H_q-1]$ in parallel}
            \STATE Compute $h_k=\left\lfloor h \times H_k /H_q\right\rfloor$ under grouped query attention.
            \FOR{each QO block $I_i$ in parallel}
                \STATE Initialize $\vO_{hi} = (0) \in \mathbb{R}^{B_q \times d_h}$, $\vell_{hi} = (0) \in \mathbb{R}^{B_q}$, and $\vm_{hi} = (-\infty) \in \mathbb{R}^{B_q}$.
                \STATE Load $\vQ_{hi}, w_{h_k}$.
                \STATE Compute the KV-block set $\mathcal{J}_{h_k}(i)$ from $w_{h_k}$.
                \FOR{$j$ descending over $\mathcal{J}_{h_k}(i)$}
                    \STATE Load $\vK_{h_kj}, \vV_{h_kj}$.
                    \STATE Compute $\vS_{hij} = \vQ_{hi} \vK_{h_kj}^\top \in \mathbb{R}^{B_q \times B_k}$ and apply causal or window mask to $\vS_{hij}$ at boundaries.
                    \STATE Compute $\tilde{m}_{hij} = \mathrm{rowmax}(\vS_{hij}) \in \mathbb{R}^{B_q}$ and $\vm_{hi}^{\mathrm{new}} = \max(\vm_{hi}, \tilde{m}_{hij})$.
                    \STATE Compute $\tilde{\vP}_{hij} = \exp(\vS_{hij} - \vm_{hi}^{\mathrm{new}}) \in \mathbb{R}^{B_q\times B_k}$.
                    \STATE Compute $\tilde{\vell}_{hij} = \mathrm{rowsum}(\tilde{\vP}_{hij}) \in \mathbb{R}^{B_q}$ and $\vell_{hi}^{\mathrm{new}} = \exp({\vm_{hi} - \vm_{hi}^{\mathrm{new}}}) \vell_{hi} + \tilde{\vell}_{hij} \in \mathbb{R}^{B_q}$.
                    \STATE Update $\vO_{hi}\leftarrow \exp({\vm_{hi} - \vm_{hi}^{\mathrm{new}}})\vO_{hi} + \tilde{\vP}_{hij} \vV_{h_kj} \in \mathbb{R}^{B_q \times d_h}$.
                    \STATE Update $\vell_{hi} \leftarrow \vell_{hi}^{\mathrm{new}}$ and $\vm_{hi} \leftarrow \vm_{hi}^{\mathrm{new}}$.
                \ENDFOR
                \STATE Update $\vO_{hi} \leftarrow \diag(\vell_{hi})^{-1} \vO_{hi}$ and $\vell_{hi} \leftarrow \vm_{hi} + \log(\vell_{hi})$.
                \STATE Store $\vO_{hi}, \vell_{hi}$.
            \ENDFOR
        \ENDFOR
        \STATE Return $\vO, \vell$.
    \end{algorithmic}
    \vspace{-0.25em}
\end{algorithm}

\subsection{Coverage and Attention Cost}
\label{sec:fwa_support_cost}

\textbf{Collective coverage.}
Under the default allocation, all KV heads share the near and sink windows, while adjacent long-range intervals provide access to the remaining causal distances.
For each query, the union of these visible-key sets covers the full causal prefix:
\begin{equation}
    \label{eq:fwa_collective_coverage}
    \bigcup_{h_k=0}^{H_k-1}\mathcal{A}_{h_k}(i_q)
    =
    \{i_k\in\{0,\ldots,N_k-1\}:\delta(i_q,i_k)\ge0\}
\end{equation}
Full causal coverage guarantees direct access to token positions, but does not imply the same head-specific interactions or outputs as FullAttn.
Each QO head applies softmax over its own visible-key set.
This union counts each accessible key once; the attention workload also depends on how many heads include that key.

\textbf{Attention cost.}
Consider the final query position $i_q=N_q-1$ and the common regime $w_{\mathrm{sink}}+w_{\mathrm{near}}\le N_k$.
The sink, near, and long-range regions are disjoint at this position, so the total number of visible keys summed across KV heads is
\begin{equation}
    \label{eq:fwa_support_counts}
    \sum_{h_k=0}^{H_k-1}\left|\mathcal{A}_{h_k}(N_q-1)\right|
    =
    H_k(w_{\mathrm{sink}}+w_{\mathrm{near}})
    +L
\end{equation}
For this query, the shared windows contribute once per KV head, while each long-range key belongs to one KV-head support set.
Dividing by $H_k$ gives the average number of visible keys per KV head.
Dense causal attention has a summed count of $H_kN_k$.
With equally sized query groups, multiplying either summed count by $H_q/H_k$ gives the actual number of QO-head query-key connections.

The final-query count also applies to each autoregressive decoding step.
Full-sequence training accumulates the support counts over all query positions; causal clipping and overlap can reduce the count for earlier queries.
The resulting work can differ across heads even when their long-range intervals have equal widths.
If $w_{\mathrm{sink}}+w_{\mathrm{near}}>N_k$, the near and sink regions already cover the causal prefix and the number of visible keys per KV head is capped by $N_k$.
For fixed $H_k$, full-sequence attention remains quadratic in sequence length, with savings arising from fewer duplicated query-key connections.
Cross-head decomposition requires $H_k>1$; when $H_k=1$, full coverage reduces to dense causal attention.

\begin{algorithm}[!t]
    \small
    \caption{\fullName{} Backward}
    \label{alg:fwa_backward}
    \vspace{-0.25em}
    \begin{algorithmic}[1]
        \REQUIRE Matrices $\vQ, \vO, \vdQ, \vdO \in \mathbb{R}^{H_q \times N_q \times d_h}$ and $\vK, \vV, \vdK, \vdV \in \mathbb{R}^{H_k \times N_k \times d_h}$. Vector $\vell \in \mathbb{R}^{H_q \times N_q}$. Windows $w$.
        \STATE Divide $\vQ, \vO, \vdQ, \vdO$ into QO blocks of size $B_q$ and $\vK, \vV, \vdK, \vdV$ into KV blocks of size $B_k$.
        \FOR{each QO head $h \in [0, \ldots, H_q-1]$ in parallel}
            \STATE Compute $h_k=\left\lfloor h \times H_k /H_q\right\rfloor$ under grouped query attention.
            \FOR{each KV block $J_j$ in parallel}
                \STATE Initialize $\vdK_{h_kj} = (0) \in \mathbb{R}^{B_k \times d_h}$ and $\vdV_{h_kj} = (0) \in \mathbb{R}^{B_k \times d_h}$.
                \STATE Load $\vK_{h_kj}, \vV_{h_kj}, w_{h_k}$.
                \STATE Compute the QO-block set $\mathcal{I}_{h_k}(j)$ from $w_{h_k}$.
                \FOR{$i$ ascending over $\mathcal{I}_{h_k}(j)$}
                    \STATE Load $\vQ_{hi}, \vO_{hi}, \vdO_{hi}$.
                    \STATE Compute $\vS_{hji} = \vK_{h_kj}\vQ_{hi}^\top \in \mathbb{R}^{B_k \times B_q}$ and apply causal or window mask to $\vS_{hji}$ at boundaries.
                    \STATE Compute $\vP_{hji} = \exp(\vS_{hji} - \vell_{hi}^\top) \in \mathbb{R}^{B_k \times B_q}$ and $\vdP_{hji} = \vV_{h_kj}\vdO_{hi}^\top \in \mathbb{R}^{B_k \times B_q}$.
                    \STATE Compute $\vdS_{hji} = \vP_{hji}\circ(\vdP_{hji} - \mathrm{rowsum}(\vO_{hi} \circ \vdO_{hi})^\top) \in \mathbb{R}^{B_k \times B_q}$.
                    \STATE Update $\vdV_{h_kj} \leftarrow \vdV_{h_kj} + \vP_{hji}\vdO_{hi}$ and $\vdK_{h_kj} \leftarrow \vdK_{h_kj} + \vdS_{hji}\vQ_{hi}$.
                    \STATE AtomicAdd $\vdQ_{hi} \leftarrow \vdQ_{hi} + \vdS_{hji}^\top\vK_{h_kj}$.
                \ENDFOR
                \STATE AtomicAdd $\vdK_{h_kj},\vdV_{h_kj}$.
            \ENDFOR
        \ENDFOR
        \STATE Return $\vdQ,\vdK,\vdV$.
    \end{algorithmic}
    \vspace{-0.25em}
\end{algorithm}

\subsection{Execution and Parallelism}
\label{sec:fwa_execution}

\textbf{Global KV-head assignment.}
\shortName{} assigns windows by global KV-head index so that tensor-parallel ranks receive complementary long-range windows.
Let $TP$ denote the number of tensor-parallel ranks, with $TP \le H_k$, and suppose the $H_k$ KV heads are evenly and contiguously sharded across them.
For local head $h_{\mathrm{local}}$ on rank $r$, we define
\begin{equation}
    \label{eq:fwa_global_head}
    h_{\mathrm{global}}
    =
    rH_k^{\mathrm{local}}+h_{\mathrm{local}}
    \qquad H_k^{\mathrm{local}}=H_k/TP
\end{equation}
The breakpoints in Equation~\ref{eq:fwa_long_range_span} are evaluated using $h_{\mathrm{global}}$ and the global $H_k$ throughout training and inference.

\textbf{Forward execution and parallelism.}
The same forward operator serves the training forward pass and inference prefill.
The QO and KV tensors are stored in head-major order, with shapes given in Algorithm~\ref{alg:fwa_forward}.
For notational simplicity, $\vQ$ is assumed to be pre-scaled by the standard factor $1/\sqrt{d_h}$, where $d_h$ is the head dimension.
We divide their token dimensions into QO blocks $I_i$ and KV blocks $J_j$ of sizes $B_q$ and $B_k$.
For QO block $I_i$, let $\mathcal{J}_{h_k}(i)$ contain the KV blocks with at least one legal pair under Equation~\ref{eq:fwa_support}.
Algorithm~\ref{alg:fwa_forward} parallelizes over QO heads and QO blocks and visits only $\mathcal{J}_{h_k}(i)$.
Blocks outside this set are skipped before QK computation; only blocks crossing a causal, window, or sequence boundary require elementwise masking.

\textbf{Backward execution and parallelism.}
For KV block $J_j$, let $\mathcal{I}_{h_k}(j)$ contain the QO blocks with at least one legal pair with $J_j$.
Algorithm~\ref{alg:fwa_backward} parallelizes over QO heads and KV blocks and traverses this inverse relation using the same visible-key set as the forward pass.
Probabilities reconstructed from the saved logsumexp $\vell$ therefore yield the exact gradient of the \shortName{} operator.
Atomic additions combine contributions to each QO gradient from multiple KV blocks and to each KV gradient from QO heads in the same group.
Under tensor parallelism, each rank applies this traversal to its local KV heads, and their global indices coordinate the sparse pattern across ranks.

\textbf{Decoding execution and parallelism.}
Autoregressive decoding is the forward case with $N_q=1$, for which $\delta(0,i_k)=N_k-1-i_k$ and the near window contains the latest KV-cache entries.
Here, $N_k$ equals the number of previously cached tokens plus one for the current token.
At each decoding step, the breakpoints are updated from this current key length, maintaining equal-width long-range windows as the sequence grows.
The visible KV blocks can be distributed across split-KV programs and combined through standard partial online-softmax states.
Split scheduling controls parallelism and load balance under the same position-defined window rule.
Across tensor-parallel ranks, global KV-head indexing preserves the allocation of complementary long-range windows, and each rank stores only its local KV heads.
KV-head sharding partitions the cache across ranks, while visiting only visible KV blocks reduces decoding-operator working memory.
During decoding in our model-level evaluations, each rank computes QKV projections for the current token and retains the full KV history of its local KV heads in HBM.
Window-aware offloading and prefetching to reduce HBM-resident KV storage remain future work.
The operator-memory measurements in Section~\ref{sec:experiments} are distinct from persistent KV-cache storage.
Configurations with $TP>H_k$ replicate logical KV heads and generally store more than a $1/TP$ fraction of the global KV cache per rank.

\section{Experiments}
\label{sec:experiments}

\textbf{Evaluation Overview.}
The experiments test whether distributing long-range access across heads preserves retrieval accuracy, reduces attention-operator cost, and maintains language-model quality across model scales.
We first isolate the role of collective coverage through a window-matched ablation, and then compare a broad set of attention mechanisms on associative recall using randomized key-value bindings.
We then benchmark the end-to-end attention-operator costs of the training forward and backward passes and autoregressive decoding for the dense reference and representative trainable sparse methods.
Finally, we study scaling behavior from 0.6B to 14B parameters and evaluate the resulting 14B models together with 32B models obtained through a separate continued-training experiment on knowledge, reasoning, and long-context retrieval benchmarks.

\textbf{Experimental Settings.}
Within each experiment, attention variants use matched model scales, depth, hidden size, data, and optimization settings, while retaining the head configuration and sparse-selection mechanism of each method.
The scaling-law and continued-training experiments are conducted on 128 NVIDIA H100 GPUs, while downstream evaluation and operator benchmarking use 8 H100 GPUs; all operator measurements use tensor parallelism with $TP=8$.
The associative-recall study includes FullAttn~\citep{vaswani2017attention}, SWA~\citep{beltagy2020longformerlongdocumenttransformer}, Seer~\citep{gao2024seerattention}, InfLLMv2~\citep{zhao2025infllmv2}, NSA~\citep{yuan2025nativesparseattentionhardwarealigned}, MoBA~\citep{lu2025moba}, DSA~\citep{deepseekai2025deepseekv32pushingfrontieropen}, and \shortName{} to cover dense, local, structured, and dynamic sparse designs.
The latency-and-memory and scaling-law comparisons focus on FullAttn, MoBA, DSA, and \shortName{}, which provide the dense reference and representative end-to-end trainable sparse mechanisms.
Complete configurations, per-query token budgets, training schedules, and evaluation protocols are provided in Appendix~\ref{sec:appendix:experiment_setup}.
Associative recall matches visible-token budgets, whereas the operator and scaling comparisons approximately match selected-attention FLOPs among the sparse methods.

\begin{table*}[!b]
    \centering
    \small
    \caption{
    \textbf{Collective Coverage with Matched Per-Head Window Widths.}
    The 1-, 2-, 4-, and 8-window layouts use matched per-head window widths and vary the number of distinct long-range windows assigned across eight KV heads. FullAttn and SWA are dense and local references and are not part of this window-matched comparison. We report associative-recall accuracy (\%) at $d_{model}=512$.
    }
    \vspace{-1.0em}
    \resizebox{\linewidth}{!}{
    \begin{tabular}{@{}lccrrrr@{}}
    \toprule
    \sc{Layout} & \sc{Window Assignment} & \sc{Long-Range Coverage}
    & \sc{1K} & \sc{2K} & \sc{4K} & \sc{8K} \\
    \midrule
    FullAttn & Full history per head & 100\% & 100 & 100 & 100 & 89.97 \\
    SWA & No long-range window & 0\% & 100 & 17.48 & 14.12 & 5.34 \\
    \midrule
    1 unique window & $1$ window $\times$ $8$ heads & 12.5\% & 100 & 53.53 & 32.99 & 21.32 \\
    2 unique windows & $2$ windows $\times$ $4$ heads & 25\% & 100 & 62.74 & 40.37 & 32.92 \\
    4 unique windows & $4$ windows $\times$ $2$ heads & 50\% & 100 & 83.91 & 61.75 & 52.32 \\
    \textbf{\shortName{}} & $8$ windows $\times$ $1$ head & 100\% & 100 & 100 & 100 & 89.73 \\
    \bottomrule
    \end{tabular}
    }
    \vspace{-2.0em}
    \label{tab:collective_coverage_ablation}
\end{table*}

\textbf{Collective Coverage with Matched Per-Head Window Widths.}
Does \shortName{} benefit from its window size, or from assigning different parts of the history to different heads?
To distinguish these factors, we match the per-head window widths and vary the number of distinct long-range windows assigned across eight KV heads.
The 1-, 2-, 4-, and 8-window layouts collectively cover 12.5\%, 25\%, 50\%, and 100\% of the long-range history, respectively; Appendix~\ref{sec:appendix:associative_recall_setup} details their construction.
Table~\ref{tab:collective_coverage_ablation} shows that 8K recall increases monotonically from 21.32\% to 32.92\%, 52.32\%, and 89.73\% as duplicated windows are replaced by complementary ones.
With full collective coverage, \shortName{} nearly matches FullAttn at 89.97\%, while the near-only SWA reference reaches 5.34\%.
These results support the benefit of distributing complementary long-range access across KV heads.

\begin{figure}[!t]
    \centering
    \includegraphics[width=\textwidth]{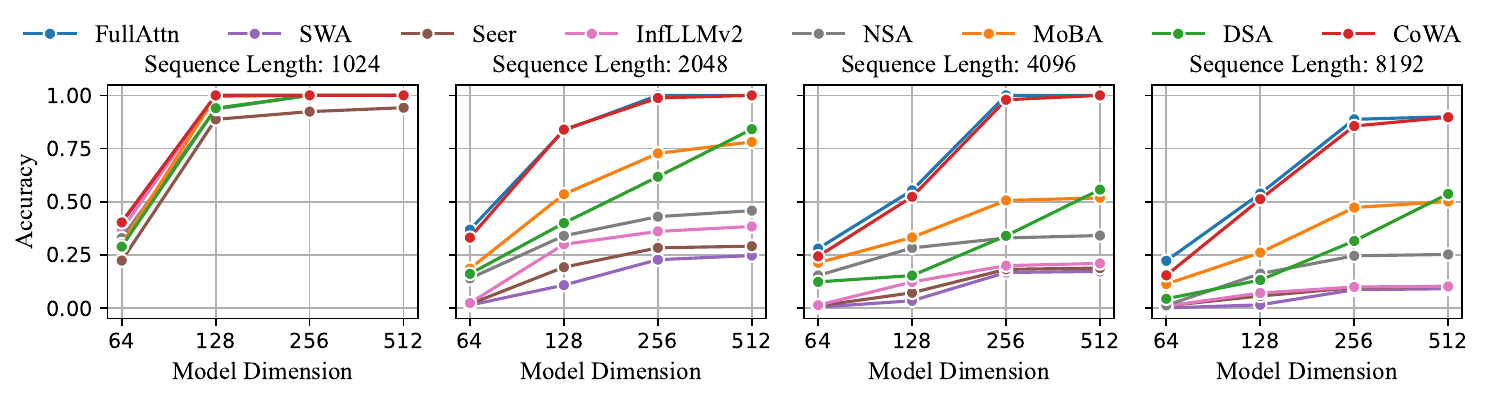}
    \vspace{-2em}
    \caption{
      \textbf{Associative Recall}.
      Accuracy with 256 key-value pairs across sequence lengths and model dimensions. At sequence lengths 1K, 2K, 4K, and 8K, all sparse methods use matched per-query token budgets of 1,024, 1,152, 1,408, and 1,920 tokens, respectively. We do not add random padding or additional noise tokens. \shortName{} closely tracks FullAttn as the context grows, whereas the other sparse patterns lose a substantial fraction of the associations.
    }
    \label{fig:ar}
    \vspace{-1.0em}
\end{figure}

\textbf{Controlled Associative Recall.}
We next use a controlled associative-recall~\citep{arora2024zoology} task to test whether each attention mechanism can recover arbitrary key-value bindings from its accessible context.
The randomized associations provide no semantic shortcut to the answer. Matching the model components outside attention helps isolate how each attention mechanism affects retrieval.
Following the setup in Appendix~\ref{sec:appendix:associative_recall_setup}, we train on 256 key-value pairs, vary the sequence length from 1,024 to 8,192 and $d_{model}$ from 64 to 512, and increase the matched per-query token budget from 1,024 to 1,920 tokens as the sequence grows.
Figure~\ref{fig:ar} shows that, at larger model dimensions, all methods approach FullAttn when the budget covers the complete 1,024-token sequence, while their behavior separates at longer sequences.
At sequence length 8,192 and $d_{model}=512$, \shortName{} reaches 89.73\% accuracy, essentially matching FullAttn at 89.97\%; DSA and MoBA reach 53.71\% and 50.12\%, NSA reaches 25.23\%, and the remaining methods stay near 10\%.
Several baselines attain higher accuracy at larger model dimensions, but increasing model dimension does not close the retrieval gap over the tested range.

\begin{figure}[!t]
    \centering
    \includegraphics[width=\textwidth]{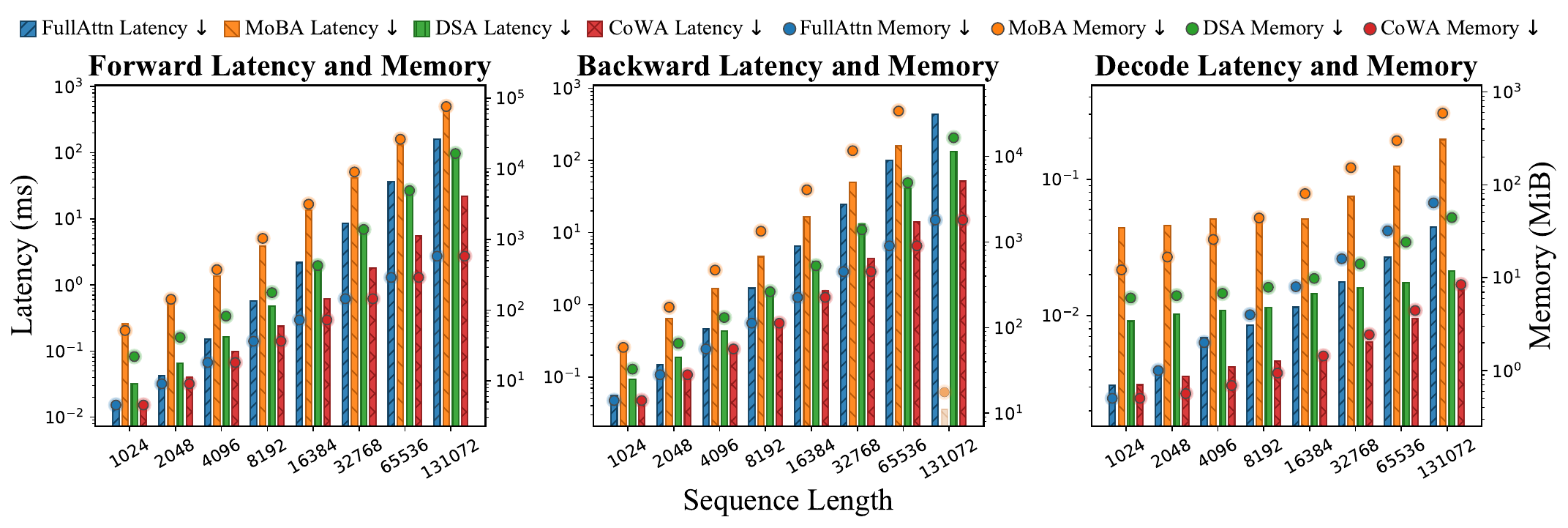}
    \vspace{-1.5em}
    \caption{
      \textbf{Operator Latency and Memory}.
      End-to-end attention-operator latency (bars, left axes) and maximum per-rank peak allocation (markers, right axes) for the training forward and backward passes and decoding with tensor parallelism $TP=8$ on 8 H100 GPUs. \shortName{} has the lowest latency and decoding-operator peak allocation among the tested methods at long sequence lengths.
    }
    \vspace{-1.5em}
    \label{fig:latency_memory}
\end{figure}

\textbf{Operator Latency and Memory.}
Having established the retrieval behavior of the attention mechanisms, we next measure the systems cost of executing the dense reference and the three representative trainable sparse designs.
The benchmark~\citep{tillet2019triton} includes sparse-pattern construction and data movement in addition to selected attention: MoBA includes block pooling, routing, TopK, rearrangement, and merging; DSA includes its lightning indexer, quantization, TopK, and sparse MLA; and \shortName{} includes its fused range and masking logic.
All measurements use the matched $TP=8$ setup described in Appendix~\ref{sec:appendix:operator_acceleration_setup}.
Figure~\ref{fig:latency_memory} shows that \shortName{}'s regular attention pattern translates into consistent gains as the context grows.
At 128K tokens, \shortName{} reduces forward and backward latency during training by $7.4\times$ and $8.6\times$, and decoding latency during inference by $3.0\times$ relative to FullAttn.
Its forward and backward peak allocations match those of the FullAttn implementation, while its 8.4 MiB decoding-operator peak allocation is $7.6\times$ smaller than FullAttn's and substantially smaller than MoBA's and DSA's.
MoBA's 128K backward pass exceeds the per-rank memory limit. MoBA and DSA incur the routing and indexing costs described above; \shortName{} derives its visible blocks directly from sequence position and global KV-head index, avoiding these auxiliary structures.

\begin{figure}[!t]
    \centering
    \includegraphics[width=\textwidth]{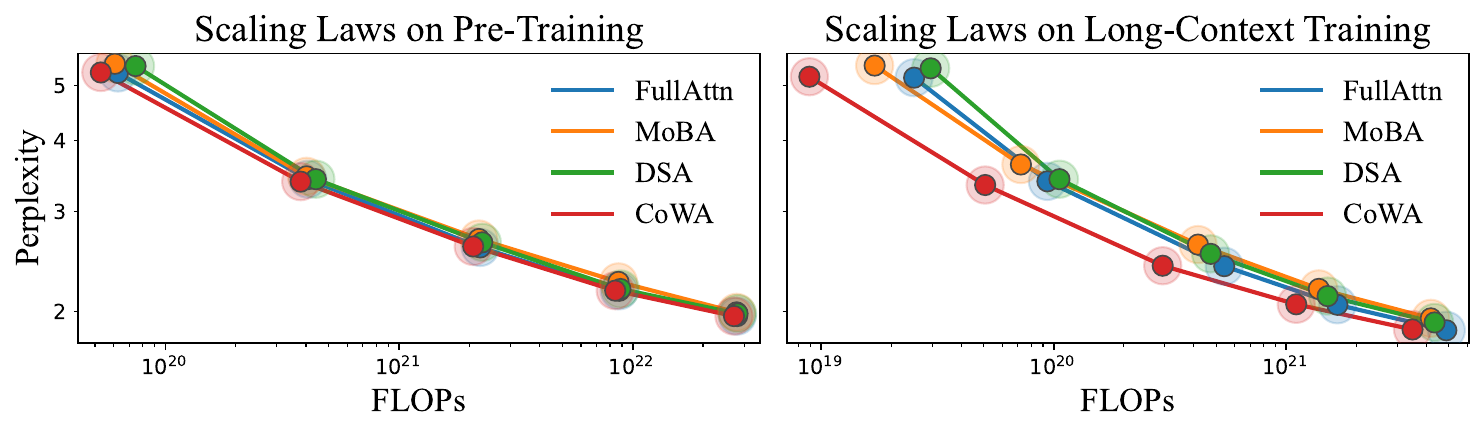}
    \vspace{-1.5em}
    \caption{
      \textbf{Scaling Laws}.
      Perplexity as a function of training FLOPs for models from 0.6B to 14B parameters during 4K pre-training (left) and 32K long-context training (right). Each marker reports final perplexity; its outer ring denotes one standard deviation over the perplexity measurements collected during training. \shortName{} closely follows FullAttn in perplexity while requiring fewer FLOPs.
    }
    \label{fig:scaling_laws_flops}
\end{figure}

\begin{table}[!t]
  \centering
  \caption{
    \textbf{Model-Level Evaluation Summary}.
    Average knowledge, reasoning, and RULER scores at 14B and 32B. RULER is evaluated at the native 32K context length and under YaRN extrapolation to 128K. Detailed per-task results and uncertainty estimates are reported in Appendix~\ref{sec:appendix:model_results}.
  }
  \vspace{-1.0em}
  \resizebox{\linewidth}{!}{
  \begin{tabular}{@{}llcccc@{}}
  \toprule
  \sc{Model Scale} & \sc{Method} & \sc{Knowledge} & \sc{Reasoning} & \sc{RULER 32K} & \sc{RULER 128K} \\
  \midrule
  14B & FullAttn & \underline{72.32} & \underline{64.46} & \textbf{89.42} & \underline{65.84} \\
  14B & MoBA & 70.58 & 59.71 & 87.17 & 61.52 \\
  14B & DSA & 70.23 & 63.28 & 87.67 & 64.05 \\
  14B & \shortName{} (ours) & \textbf{72.70} & \textbf{64.87} & \underline{89.13} & \textbf{66.60} \\
  \midrule
  32B & FullAttn & \underline{75.62} & \textbf{75.67} & \textbf{92.70} & \textbf{82.03} \\
  32B & \shortName{} (ours) & \textbf{76.07} & \underline{75.53} & \underline{92.58} & \underline{81.78} \\
  \bottomrule
  \end{tabular}
  }
  \vspace{-2.0em}
  \label{table:model_level_summary}
\end{table}

\textbf{Scaling Across Model Sizes.}
We next test the quality-compute trade-off during language-model training across model sizes~\citep{kaplan2020scalinglawsneurallanguage, xiong2023effectivelongcontextscalingfoundation}.
We study five model scales from 0.6B to 14B under matched pre-training and long-context training configurations.
Figure~\ref{fig:scaling_laws_flops} reports perplexity against total training FLOPs, including the sequence-dependent attention and sparse-selection work.
Across both stages, \shortName{} closely tracks FullAttn in perplexity while requiring less computation, giving the best overall perplexity-FLOPs trade-off among the sparse variants.
At 14B, \shortName{} matches FullAttn's pre-training perplexity and differs by less than 0.01 after long-context training, while reducing total training FLOPs by 3.1\% and 28.5\%, respectively.
The FLOP reduction is largest in the 32K stage, where the cost of dense attention grows with sequence length, while \shortName{} uses a per-QO-head token budget of 4,992 without introducing router or indexer FLOPs.
Detailed model architectures, token budgets, FLOP accounting, and optimization settings are provided in Appendix~\ref{sec:appendix:scaling_laws_setup}.

\textbf{Model-Level Evaluation.}
We finally test whether the preceding trends transfer to complete language models.
We benchmark all four 14B attention variants obtained from the scaling-law study and extend the FullAttn-\shortName{} comparison to the 32B models obtained through the separate continued-training experiment; the latter are included only in this evaluation and not in the scaling-law analysis.
Table~\ref{table:model_level_summary} shows that \shortName{} remains comparable to FullAttn in average knowledge, reasoning, native 32K retrieval, and YaRN-extrapolated 128K retrieval at both scales.
At 14B, \shortName{} scores 72.70 and 64.87 on knowledge and reasoning versus 72.32 and 64.46 for FullAttn.
At 32B, the corresponding averages are 76.07 and 75.53 versus 75.62 and 75.67.
\shortName{} remains within 0.3 points of FullAttn at native 32K for both model scales.
Under YaRN~\citep{peng2026yarn} extrapolation to 128K, \shortName{} reaches 66.60 versus 65.84 for FullAttn at 14B and 81.78 versus 82.03 at 32B.
Bold and underlining mark the best and second-best results, respectively, within each column and model scale.
Detailed per-task results and uncertainty estimates are reported in Appendix~\ref{sec:appendix:model_results}.
The 14B results connect lower training FLOPs to FullAttn-level average benchmark performance, while the 32B comparison shows that comparable quality can also be retained after continued training.

\section{Related Work}
\label{sec:related_work}

\shortName{} defines sparse access directly from position during training and inference, replacing duplicated access to the causal history with complementary long-range windows across KV heads; we relate this design to sparse attention, head specialization and routing, hardware-aware sparse execution, and KV-cache management and distributed inference.

\textbf{Sparse attention.}
Fixed sparse attention uses local windows, structured blocks, or global tokens~\citep{child2019generating,beltagy2020longformerlongdocumenttransformer,zaheer2020big}.
Sliding-window attention allows regular kernel execution, but restricts every head to the same finite horizon~\citep{fu2025slidingwindowattentiontraining}.
Dynamic methods select tokens or blocks using retrieval scores, importance estimates, or learned routers~\citep{tang2024quest,li2024snapkv,zhang2023h2o,xiao2024infllm,desai2024hashattention,zhang2025damdynamicattentionmask,gao2024seerattention,yuan2025nativesparseattentionhardwarealigned,lu2025moba,lai2025flexprefill}.
\shortName{} combines shared near and sink windows with complementary long-range windows across KV heads, retaining full causal coverage through a position-defined rule without content-dependent selection.

\textbf{Head specialization.}
Attention heads exhibit distinct local and long-range roles~\citep{guo2024activedormantah,gu2024whenas,barbero2025whydl,xiao2024duoattentionel,sandovalsegura2025identifyingae}.
SPAttention distributes causal-distance regions across MHA heads while preserving aggregate causal connectivity~\citep{zhao2025makingheadcountsparse}.\footnote{SPAttention's published exclusive bands differ from its released shared-local implementation \href{https://github.com/Harry-Miral/MEHC}{\nolinkurl{MEHC}}, which does not provide the long-context GQA, decoding, or tensor-parallel execution studied here.}
We systematically evaluate collective coverage as a design principle for sparse attention: an attention layer retains access to the full causal history collectively across heads, without requiring every head to attend to every position.
\shortName{}'s complementary KV-head windows provide a simple instantiation of this principle under GQA.
This allocation of visible keys differs from token-dependent gating of head outputs~\citep{fu2026attentionsinkforgesnative}.
Beyond the particular distance partition, we study this principle through controlled coverage ablations, evaluations of language-model quality and long-range retrieval, and benchmarks of tensor-parallel training and inference operators.

\textbf{Hardware-aware sparse execution.}
FlashAttention reduces memory traffic through tiled execution and online softmax while preserving dense attention~\citep{dao2022flashattention,shah2024flashattention3}; FlexAttention and FlashInfer improve kernel programmability and serving efficiency~\citep{dong2024flexattentionprogrammingmodel,ye2025flashinferefficientcustomizableattention}.
\shortName{} uses tiled block traversal but derives visible block ranges directly from position and global KV-head index, avoiding separate routing, indexing, and materialized selection state.
Its position-defined sparsity is part of the trained attention architecture rather than an inference-only approximation.

\textbf{KV-cache management and distributed inference.}
KV-cache systems reduce inference cost by pruning, compressing, retrieving, or offloading cached states~\citep{zhang2023h2o,li2024snapkv,tang2024quest,xiao2024infllm,liu2024clusterkv,huang2025nosanativeoffloadablesparse}.
\shortName{} instead changes the trained attention architecture and aligns complementary windows with tensor-parallel KV-head sharding; cache compression and offloading remain orthogonal.

\section{Conclusion}

We introduced \fullName{} (\shortName{}), a structured architecture that distributes access to the causal history across KV heads using one position-defined window rule for training and inference. \shortName{} shares near and sink windows while assigning complementary long-range windows to different heads, making full causal coverage a collective property of the head ensemble. This design reduces duplicated long-range access without a learned router or indexer. Global KV-head indexing aligns the window assignment with tensor parallelism and per-rank KV-cache partitioning.

The window-matched ablation shows that complementary long-range windows improve recall compared with duplicated windows at the same per-head widths. The controlled associative recall further shows that \shortName{} closely tracks FullAttn as context length grows. The tensor-parallel operator benchmarks show lower attention latency and peak decoding-operator memory. The scaling-law experiments from 0.6B to 14B parameters show perplexity comparable to FullAttn with lower training FLOPs, and the model-level evaluations at 14B and 32B show comparable knowledge, reasoning, and long-context retrieval scores. These results indicate that distributing long-range access across KV heads can retain the evaluated capabilities of FullAttn while reducing attention computation.



\bibliography{biblio}
\bibliographystyle{iclr2027_conference}

\appendix
\newpage
\section{Experiment Settings}
\label{sec:appendix:experiment_setup}

Within each experiment, attention variants use matched model scales, depth, hidden size, data, and optimization settings; attention-specific head configurations and sparse hyperparameters follow each method.
Our software stack uses NVIDIA PyTorch container images~\citep{nv2022pytorch} and the Transformers framework~\citep{wolf-etal-2020-transformers}.
The scaling-law~\citep{kaplan2020scalinglawsneurallanguage, xiong2023effectivelongcontextscalingfoundation} and continued-training experiments use Megatron-LM~\citep{megatron-lm} for distributed training on 128 NVIDIA H100 GPUs.
We use the EleutherAI LM Evaluation Harness~\citep{eval-harness} for benchmark evaluation on 8 NVIDIA H100 GPUs. Operator benchmarking uses 8 H100 GPUs with tensor parallelism $TP=8$.

\subsection{Associative Recall Setup}
\label{sec:appendix:associative_recall_setup}

Following prior work~\citep{arora2024zoology}, we evaluate associative recall with 256 key-value pairs. We consider sequence lengths of 1,024, 2,048, 4,096, and 8,192, and use the standard model dimensions $d_{model}\in\{64,128,256,512\}$. The repeated key-value content is placed contiguously at the beginning of each sequence, occupying the first $2n_{\mathrm{kv}}n_{\mathrm{pass}}$ tokens, where $n_{\mathrm{kv}}=256$ and $n_{\mathrm{pass}}$ is the number of passes. We then sample $n_{\mathrm{kv}}$ query positions without replacement from the remaining positions according to $p(t)\propto t^{a-1}$, where $t$ is the position within this suffix. We use the default $a=0.01$, so queries occur more frequently near the beginning of the suffix and become progressively sparser toward the end. We do not insert random padding or additional noise tokens. This setup tests recall without adding a separate task of filtering random padding or noise.

The dataset contains 250K training examples and 1K test examples, and all models are trained for 100 epochs. At the four sequence lengths, every sparse attention variant uses matched per-query token budgets of 1,024, 1,152, 1,408, and 1,920 tokens, respectively. For \shortName{}, we use \(w_{\mathrm{sink}}=1\) and \(w_{\mathrm{near}}=1{,}023\), giving a shared width of 1,024 tokens; the complementary long-range windows produce the stated effective per-QO-head token budgets after overlap and sequence-boundary clipping.

\paragraph{Collective-coverage layouts.}
We index the eight equal-width long-range regions from the current query toward the beginning of the sequence: $W_1$ is adjacent to the near window and $W_8$ is adjacent to the sink window. Reading from the earliest token to the current query, the union of the \shortName{} heads is therefore $\mathrm{sink}\mid W_8\mid W_7\mid\cdots\mid W_2\mid W_1\mid\mathrm{near}$. All \shortName{} heads share the sink and near windows, and each KV head receives one distinct $W_j$. In the collective-coverage ablation, SWA receives no sink or long-range window and attends only through its near window; it is included as a local reference rather than a work-matched variant. For an $m$-window layout, we repeat the $m$ windows closest to the near window, $W_1,\ldots,W_m$, across the eight KV heads. Thus, 1, 2, or 4 distinct long-range windows are each repeated across 8, 4, or 2 KV heads, respectively. These layouts and \shortName{} use matched per-head window widths while increasing collective coverage.

\subsection{Operator Acceleration Setup}
\label{sec:appendix:operator_acceleration_setup}

We separate the cost of sparse attention into computation over selected tokens and the work required to determine the sparse pattern. Let $n$ denote sequence length, $h$ the number of QO heads, $d_{qk}$ and $d_v$ the query/key and value head dimensions, $B$ the MoBA block size, and $w$ the per-query token budget. Thus, $w$ denotes the number of selected tokens after expanding MoBA blocks, the number of top-ranked tokens selected by DSA, and the per-QO-head token budget for \shortName{}. For DSA, $h_I$ and $d_I$ denote the number and dimension of lightning-indexer heads. Table~\ref{tab:attention_comparison_intro} reports the resulting asymptotic costs. Projection layers and other sequence-linear Transformer operations are shared model costs and are omitted from this table.

FullAttn computes QK and PV over all token pairs. MoBA~\citep{lu2025moba} reduces the main attention to selected blocks, but first mean-pools block keys and evaluates every query against the $n/B$ block representatives. DSA~\citep{deepseekai2025deepseekv32pushingfrontieropen} performs token-level sparse MLA after a lightning indexer scores all query-key pairs and generates TopK indices. \shortName{} requires no separate routing or indexing stage; its fused range calculation directly restricts QK and PV to the retained tokens.

Consequently, methods with the same selected-token budget can have substantially different end-to-end costs once routing and indexing are included.

\begin{table}[H]
    \centering
    \caption{
    \textbf{Comparison of Different Attention Variants}.
    The sequence-dependent attention and sparse-selection costs. The auxiliary state column describes routing or indexing state beyond the QKV tensors and attention outputs.
    }
    \label{tab:attention_comparison_intro}
    \resizebox{\linewidth}{!}{
    \begin{tabular}{@{}lccc@{}}
    \toprule
    \sc{Mechanism} & \sc{Selected Attention} & \sc{Selection Overhead} & \sc{Auxiliary State} \\
    \midrule
    FullAttn & $O\!\left(n^2 h(d_{qk}+d_v)\right)$ & - & - \\
    MoBA & $O\!\left(nwh(d_{qk}+d_v)\right)$ & $O\!\left(n^2 h d_{qk}/B\right)$ & $O\!\left(n^2h/B\right)$ \\
    DSA & $O\!\left(nwh(d_{qk}+d_v)\right)$ & $O\!\left(n^2h_Id_I\right)$ & $O\!\left(n^2h_I+nw\right)$ \\
    \textbf{\shortName{}} & $O\!\left(nwh(d_{qk}+d_v)\right)$ & $O(1)$ per QO block & $O(H_k)$ \\
    \bottomrule
    \end{tabular}
    }
\end{table}

\footnotetext[1]{The FullAttn implementation is available at \href{https://github.com/Dao-AILab/flash-attention}{\nolinkurl{github.com/Dao-AILab/flash-attention}}.}
\footnotetext[2]{The MoBA implementation is available at \href{https://github.com/MoonshotAI/MoBA}{\nolinkurl{https://github.com/MoonshotAI/MoBA}}.}
\footnotetext[3]{The DSA indexer implementation is available at \href{https://github.com/deepseek-ai/DeepGEMM}{\nolinkurl{github.com/deepseek-ai/DeepGEMM}}.}
\footnotetext[4]{The DSA implementation is available at \href{https://github.com/Dao-AILab/flash-attention/tree/main/flash_attn/cute}{\nolinkurl{github.com/Dao-AILab/flash-attention/flash_attn/cute}}.}

We benchmark~\citep{tillet2019triton} causal attention at sequence lengths from 1,024 to 131,072 on 8 NVIDIA H100 GPUs (SM90) with tensor parallelism $TP=8$. All methods use batch size one and the same tensor-parallel configuration. Unless a configuration runs out of memory, latency is averaged over 100 iterations after 25 warm-up iterations. We report the synchronized wall-clock time of the tensor-parallel group, and peak memory is the maximum per-rank allocation across the eight ranks. Table~\ref{tab:latency_memory_configs} summarizes the operator shapes, precision, and sparse-selection mechanism used by each method, while Table~\ref{tab:latency_memory_budgets} reports the exact retained-token budgets passed at every sequence length.

The \textbf{Per-Selected-Token FLOPs} (PST FLOPs) column reports the forward QK and PV FLOPs per query contributed by one selected key token, summed over all QO heads. Counting one multiply-accumulate as two FLOPs, it is $2h(d_{qk}+d_v)$. This quantity excludes router, indexer, TopK, and other sparse-pattern construction costs; it is used only to approximately match the forward selected-attention workload. MoBA budgets are aligned to its 128-token blocks, while DSA uses approximately one quarter of the \shortName{} budget because its PST FLOPs are approximately four times as large. Total training FLOPs separately count QK recomputation and the value-side backward operations. FullAttn always attends to the complete causal context.

The reported measurements of the training forward and backward passes and inference-time decoding are operator-level end-to-end timings rather than isolated sparse-attention kernel timings. In particular, MoBA includes block representative construction, router logits, TopK, index construction, data rearrangement, and output merging. DSA includes BF16 indexer inputs, FP8 quantization, indexer logits, TopK/indices, and sparse MLA; decoding quantizes and appends only the newly generated Q/KV token before index selection and sparse attention. \shortName{} includes its fused window-range and masking logic. Model-level QKV/output projections, MLP layers, communication, and optimizer updates are outside the scope of this module benchmark. Peak memory is measured over the same end-to-end operator path and does not represent persistent serving-time KV-cache size.

\begin{table}[H]
    \centering
    \small
    \caption{
    \textbf{Latency and Memory Benchmark Configurations}.
    The operator shapes and sparse-selection mechanisms used for the running-time curves. All measurements use tensor parallelism $TP=8$ on 8 H100 GPUs.
    }
    \label{tab:latency_memory_configs}
    \resizebox{\linewidth}{!}{
    \begin{tabular}{@{}lccccccccc@{}}
    \toprule
    \sc{Algo} & warmups & runs & batch & $n_h$ & $n_{h_{kv}}$ & $d_{qk}$ & $d_v$ & selector & precision \\
    \midrule
    FullAttn\footnotemark[1] & 25 & 100 & 1 & 64 & 8 & 128 & 128 & - & BF16 \\
    MoBA\footnotemark[2] & 25 & 100 & 1 & 64 & 64 & 128 & 128 & $B=128$ & BF16 \\
    DSA\footnotemark[3]\textsuperscript{,}\footnotemark[4] & 25 & 100 & 1 & 64 & 1 & 576 & 512 & FP8 indexer & BF16/FP8 \\
    \shortName{} & 25 & 100 & 1 & 64 & 8 & 128 & 128 & fused window & BF16 \\
    \bottomrule
    \end{tabular}
    }
\end{table}

\begin{table}[H]
    \centering
    \small
    \caption{
    \textbf{Per-Query Token Budgets}.
    The token budgets used at every benchmark sequence length.
    }
    \label{tab:latency_memory_budgets}
    \resizebox{\linewidth}{!}{
    \begin{tabular}{@{}lcccccccccc@{}}
    \toprule
    \sc{Algo} & \sc{PST FLOPs} & 1K & 2K & 4K & 8K & 16K & 32K & 64K & 128K \\
    \midrule
    FullAttn & 32768 & 1024 & 2048 & 4096 & 8192 & 16384 & 32768 & 65536 & 131072 \\
    MoBA & 32768 & 1024 & 1024 & 1536 & 2048 & 3072 & 5120 & 9216 & 17408 \\
    DSA & 139264 & 256 & 288 & 352 & 480 & 736 & 1248 & 2272 & 4320 \\
    \shortName{} & 32768 & 1024 & 1152 & 1408 & 1920 & 2944 & 4992 & 9088 & 17280 \\
    \bottomrule
    \end{tabular}
    }
\end{table}

\begin{table}[H]
    \centering
    \small
    \caption{
    \textbf{Latency and Memory Benchmark Results}.
    End-to-end latency (ms) and peak memory (MiB) for the training forward and backward passes and inference-time decoding at every sequence length. Latency is the synchronized tensor-parallel group latency, and memory is the maximum per-rank peak allocation.
    }
    \label{tab:latency_memory_results}
    \resizebox{\linewidth}{!}{
    \begin{tabular}{@{}lcccccccccc@{}}
    \toprule
    \sc{Algo} & SeqLen & Fwd Lat. & Bwd Lat. & Dec Lat. & Fwd Mem. & Bwd Mem. & Dec Mem. \\
    \midrule
    FullAttn & 1,024 & 0.015 & 0.055 & 0.003 & 4.531 & 14.094 & 0.504 \\
    FullAttn & 2,048 & 0.043 & 0.149 & 0.004 & 9.063 & 28.188 & 1.004 \\
    FullAttn & 4,096 & 0.153 & 0.468 & 0.007 & 18.125 & 56.375 & 2.004 \\
    FullAttn & 8,192 & 0.572 & 1.699 & 0.009 & 36.250 & 112.750 & 4.004 \\
    FullAttn & 16,384 & 2.179 & 6.455 & 0.012 & 72.500 & 225.500 & 8.004 \\
    FullAttn & 32,768 & 8.570 & 24.449 & 0.018 & 145.000 & 451.000 & 16.004 \\
    FullAttn & 65,536 & 36.431 & 99.664 & 0.027 & 290.000 & 902.000 & 32.004 \\
    FullAttn & 131,072 & 159.987 & 441.230 & 0.045 & 580.000 & 1,804.001 & 64.004 \\
    \midrule
    MoBA & 1,024 & 0.259 & 0.231 & 0.044 & 51.425 & 58.825 & 12.162 \\
    MoBA & 2,048 & 0.563 & 0.639 & 0.046 & 142.287 & 173.275 & 16.712 \\
    MoBA & 4,096 & 1.348 & 1.668 & 0.051 & 373.875 & 470.188 & 25.800 \\
    MoBA & 8,192 & 3.833 & 4.638 & 0.051 & 1,035.975 & 1,335.475 & 43.987 \\
    MoBA & 16,384 & 13.577 & 16.429 & 0.051 & 3,156.900 & 4,069.559 & 80.438 \\
    MoBA & 32,768 & 41.543 & 50.268 & 0.075 & 9,029.144 & 11,639.466 & 153.125 \\
    MoBA & 65,536 & 131.851 & 159.542 & 0.124 & 26,236.976 & 33,822.076 & 298.637 \\
    MoBA & 131,072 & 418.473 & OOM & 0.196 & 76,239.664 & OOM & 589.650 \\
    \midrule
    DSA & 1,024 & 0.032 & 0.093 & 0.009 & 21.988 & 32.725 & 6.062 \\
    DSA & 2,048 & 0.065 & 0.187 & 0.010 & 41.100 & 65.562 & 6.412 \\
    DSA & 4,096 & 0.164 & 0.431 & 0.011 & 82.200 & 131.125 & 6.800 \\
    DSA & 8,192 & 0.481 & 1.324 & 0.011 & 176.887 & 262.750 & 7.900 \\
    DSA & 16,384 & 1.668 & 3.955 & 0.015 & 427.100 & 527.500 & 9.825 \\
    DSA & 32,768 & 6.231 & 13.199 & 0.016 & 1,388.225 & 1,388.225 & 14.075 \\
    DSA & 65,536 & 24.616 & 40.914 & 0.017 & 4,924.562 & 4,924.562 & 24.150 \\
    DSA & 131,072 & 93.671 & 133.231 & 0.021 & 16,479.996 & 16,479.996 & 44.388 \\
    \midrule
    \shortName{} & 1,024 & 0.015 & 0.041 & 0.003 & 4.531 & 14.094 & 0.504 \\
    \shortName{} & 2,048 & 0.040 & 0.106 & 0.003 & 9.063 & 28.188 & 0.566 \\
    \shortName{} & 4,096 & 0.098 & 0.253 & 0.004 & 18.125 & 56.375 & 0.691 \\
    \shortName{} & 8,192 & 0.237 & 0.600 & 0.004 & 36.250 & 112.750 & 0.941 \\
    \shortName{} & 16,384 & 0.616 & 1.564 & 0.005 & 72.500 & 225.500 & 1.441 \\
    \shortName{} & 32,768 & 1.807 & 4.358 & 0.006 & 145.000 & 451.000 & 2.441 \\
    \shortName{} & 65,536 & 5.573 & 14.150 & 0.009 & 290.000 & 902.000 & 4.441 \\
    \shortName{} & 131,072 & 21.758 & 51.509 & 0.015 & 580.000 & 1,804.001 & 8.441 \\
    \bottomrule
    \end{tabular}
    }
\end{table}

\subsection{Scaling Laws Setup}
\label{sec:appendix:scaling_laws_setup}

The scaling-law~\citep{kaplan2020scalinglawsneurallanguage, xiong2023effectivelongcontextscalingfoundation} experiments use SmolLMCorpus~\citep{benallal2024smollmcorpus}, the Qwen3 tokenizer~\citep{qwen32025}, the AdamW optimizer~\citep{Loshchilov2017FixingWD}, the WSD learning-rate schedule~\citep{hägele2024scalinglawscomputeoptimaltraining}, and the compute-optimal scaling principles~\citep{li2025predictablescalei,hoffmann2022empirical}. We use a fixed random seed of 42 for all scaling-law runs.
We summarize the meaning of the columns in Table~\ref{tab:scaling_laws_configurations} and clarify which hyperparameters are used by each attention variant.

\begin{itemize}
    \item \textbf{Params}: target model scale. Parameter counts are matched approximately because the attention projections differ across variants.
    \item \textbf{PT/LCT Tok}: total numbers of tokens used in pre-training (PT) and long-context training (LCT), respectively. Each table entry reports the two values in the order \texttt{PT, LCT}.
    \item \textbf{Batch}: tokens per optimization step.
    \item \textbf{PT/LCT LR}: learning rates used during PT and LCT, respectively. Each table entry reports the two values in the order \texttt{PT, LCT}.
    \item $n_{layers}$: number of Transformer layers.
    \item $d_{model}$: model hidden size.
    \item $n_h$: number of attention heads.
    \item $n_{h_{kv}}$: number of KV heads.
    \item $d_{qk}$ and $d_v$: per-head dimensions of the query/key and value representations, respectively.
    \item \textbf{Per-Selected-Token FLOPs} (PST FLOPs): forward QK and PV FLOPs per query contributed by each selected key token, summed over all QO heads, as defined in Appendix~\ref{sec:appendix:operator_acceleration_setup}.
    \item \textbf{PT Bgt} and \textbf{LCT Bgt}: the per-query token budgets used during pre-training and long-context training, respectively.
\end{itemize}

For models from 0.6B to 14B, training proceeds in two stages. PT uses a sequence length of 4,096 and a WSD learning-rate schedule~\citep{hägele2024scalinglawscomputeoptimaltraining} whose peak learning rate is reported in the PT entry of Table~\ref{tab:scaling_laws_configurations}. The decay phase ends at the corresponding LCT learning rate, which is $0.1\times$ the PT peak learning rate. LCT initializes from the final PT checkpoint, increases the sequence length to 32,768, and uses the reported constant LCT learning rate.

\begin{itemize}
    \item \textbf{FullAttn}: standard causal scaled dot-product attention over the complete pre-training or long-context sequence.
    \item \textbf{MoBA}: block-sparse attention with one KV head per QO head and retained-token budgets of 1,536 during pre-training and 5,120 during long-context training.
    \item \textbf{DSA}: sparse latent attention with one KV head, $d_{qk}=576$, and $d_v=512$. Since each selected token requires approximately four times the forward QK and PV FLOPs of MoBA or \shortName{}, we use budgets of 352 and 1,248.
    \item \textbf{\shortName{}}: grouped-query sparse attention with eight KV heads and per-QO-head token budgets of 1,408 during pre-training and 4,992 during long-context training.
\end{itemize}

We retain each method's native head configuration: FullAttn and \shortName{} use eight KV heads, MoBA uses one KV head per QO head, and DSA uses a single latent KV head. For the sparse variants, we approximately match forward selected-attention FLOPs rather than head counts or raw token budgets. The token budgets differ because of each method's per-token cost and block or token-selection granularity. This matching criterion excludes method-specific routing, indexing, and top-$k$ overheads and does not require identical end-to-end training FLOPs. FullAttn attends to the complete context and serves as the dense reference.

The scaling-law plots use cumulative training FLOPs reported by Megatron-LM and therefore cover the complete model computation. To expose the attention-specific contribution to this statistic, we additionally measure the exact number of legal attention connections executed under each training configuration rather than assigning the nominal final-query budget to every query. Let $\bar{s}$ denote the resulting mean number of selected keys per query and QO head. Attention backward recomputes QK scores, so the sequence-dependent core-attention cost per token and layer is $8\bar{s}hd_{qk}+6\bar{s}hd_v$, counting one multiply-accumulate as two FLOPs. FullAttn has $\bar{s}=(n+1)/2$ over the causal triangle. For \shortName{}, averaging the exact support sizes over all query positions and QO heads gives $\bar{s}=1{,}040.17$ during 4K pre-training and $\bar{s}=2{,}930.08$ during 32K long-context training. MoBA and DSA use their causally clipped selected supports, with their routing and indexing costs included in the reported totals.

\begin{table}[H]
    \centering
    \small
    \caption{
    \textbf{Self-Attention Variants Scaling Laws Configurations}.
    The model and hyperparameter configurations used in our self-attention variants scaling laws experiments.
    }
    \label{tab:scaling_laws_configurations}
    \resizebox{\linewidth}{!}
    {
    \begin{tabular}{@{}lccccccccccccc@{}}
    \toprule
    \sc{Algos} & \sc{Params} & \sc{PT/LCT Tok} & \sc{Batch} & \sc{PT/LCT LR} & $n_{layers}$ & $d_{model}$ & $n_h$ & $n_{h_{kv}}$ & $d_{qk}$ & $d_v$ & \sc{PST FLOPs} & \sc{PT Bgt} & \sc{LCT Bgt} \\
    \midrule
    FullAttn & $\approx$ 0.6B & 12B,1.5B & 0.256M & 1e-3,1e-4 & 28 & 1024 & 16 & 8 & 128 & 128 & 8192 & 4096 & 32768 \\
    MoBA & $\approx$ 0.6B & 12B,1.5B & 0.256M & 1e-3,1e-4 & 28 & 1024 & 16 & 16 & 128 & 128 & 8192 & 1536 & 5120 \\
    DSA & $\approx$ 0.6B & 12B,1.5B & 0.256M & 1e-3,1e-4 & 28 & 1024 & 16 & 1 & 576 & 512 & 34816 & 352 & 1248 \\
    \shortName{} & $\approx$ 0.6B & 12B,1.5B & 0.256M & 1e-3,1e-4 & 28 & 1024 & 16 & 8 & 128 & 128 & 8192 & 1408 & 4992 \\
    \midrule
    FullAttn & $\approx$ 1.7B & 34B,4B & 0.512M & 8e-4,8e-5 & 28 & 2048 & 16 & 8 & 128 & 128 & 8192 & 4096 & 32768 \\
    MoBA & $\approx$ 1.7B & 34B,4B & 0.512M & 8e-4,8e-5 & 28 & 2048 & 16 & 16 & 128 & 128 & 8192 & 1536 & 5120 \\
    DSA & $\approx$ 1.7B & 34B,4B & 0.512M & 8e-4,8e-5 & 28 & 2048 & 16 & 1 & 576 & 512 & 34816 & 352 & 1248 \\
    \shortName{} & $\approx$ 1.7B & 34B,4B & 0.512M & 8e-4,8e-5 & 28 & 2048 & 16 & 8 & 128 & 128 & 8192 & 1408 & 4992 \\
    \midrule
    FullAttn & $\approx$ 4B & 80B,10B & 1M & 6e-4,6e-5 & 32 & 2560 & 32 & 8 & 128 & 128 & 16384 & 4096 & 32768 \\
    MoBA & $\approx$ 4B & 80B,10B & 1M & 6e-4,6e-5 & 32 & 2560 & 32 & 32 & 128 & 128 & 16384 & 1536 & 5120 \\
    DSA & $\approx$ 4B & 80B,10B & 1M & 6e-4,6e-5 & 32 & 2560 & 32 & 1 & 576 & 512 & 69632 & 352 & 1248 \\
    \shortName{} & $\approx$ 4B & 80B,10B & 1M & 6e-4,6e-5 & 32 & 2560 & 32 & 8 & 128 & 128 & 16384 & 1408 & 4992 \\
    \midrule
    FullAttn & $\approx$ 8B & 164B,20B & 1.6M & 4e-4,4e-5 & 36 & 4096 & 32 & 8 & 128 & 128 & 16384 & 4096 & 32768 \\
    MoBA & $\approx$ 8B & 164B,20B & 1.6M & 4e-4,4e-5 & 36 & 4096 & 32 & 32 & 128 & 128 & 16384 & 1536 & 5120 \\
    DSA & $\approx$ 8B & 164B,20B & 1.6M & 4e-4,4e-5 & 36 & 4096 & 32 & 1 & 576 & 512 & 69632 & 352 & 1248 \\
    \shortName{} & $\approx$ 8B & 164B,20B & 1.6M & 4e-4,4e-5 & 36 & 4096 & 32 & 8 & 128 & 128 & 16384 & 1408 & 4992 \\
    \midrule
    FullAttn & $\approx$ 14B & 296B,36B & 2M & 3e-4,3e-5 & 40 & 5120 & 40 & 8 & 128 & 128 & 20480 & 4096 & 32768 \\
    MoBA & $\approx$ 14B & 296B,36B & 2M & 3e-4,3e-5 & 40 & 5120 & 40 & 40 & 128 & 128 & 20480 & 1536 & 5120 \\
    DSA & $\approx$ 14B & 296B,36B & 2M & 3e-4,3e-5 & 40 & 5120 & 40 & 1 & 576 & 512 & 87040 & 352 & 1248 \\
    \shortName{} & $\approx$ 14B & 296B,36B & 2M & 3e-4,3e-5 & 40 & 5120 & 40 & 8 & 128 & 128 & 20480 & 1408 & 4992 \\
    \bottomrule
    \end{tabular}
    }
\end{table}

\subsection{Sparse Adaptation via Continued Training}
\label{sec:appendix:sparse_adaptation_setup}

We study sparse adaptation at 32B through long-context continued training, using FullAttn as the dense reference.
Both models are trained at a sequence length of 32,768 for 64B tokens with a batch size of 4M tokens and a constant learning rate of $1\times10^{-5}$.
FullAttn uses the full 32,768-token causal context, and \shortName{} uses a per-QO-head token budget of 4,992.
Both configurations use 64 layers, a hidden size of 5,120, 64 QO heads, 8 KV heads, and per-head dimensions $d_{qk}=d_v=128$.

\subsection{Benchmark Evaluation}
\label{sec:appendix:benchmark_setup}

We evaluate two groups of models.
At 14B, we compare all four attention variants obtained from the scaling-law study in Appendix~\ref{sec:appendix:scaling_laws_setup}.
At 32B, we compare the FullAttn and \shortName{} models obtained from the continued-training setup in Appendix~\ref{sec:appendix:sparse_adaptation_setup}.
For each task, we report the mean and standard deviation over five evaluation runs with seeds 0, 42, 233, 666, and 1234.
The standard benchmark suite covers knowledge, reasoning, and retrieval
through MMLU~\citep{hendrycks2021measuring,lyu2024probabilitiesunveilingmisalignmentevaluating}, MMLU-Pro~\citep{wang2024mmluprorobustchallengingmultitask}, BBH~\citep{suzgun2023challenging}, HellaSwag~\citep{zellers2019hellaswag}, OBQA~\citep{mihaylov2018can}, WinoGrande~\citep{sakaguchi2021winogrande}, PIQA~\citep{bisk2020piqa}, GSM8K~\citep{cobbe2021training}, Hendrycks-Math~\citep{hendrycks2021measuringmathematicalproblemsolving}, ARC-C~\citep{clark2018think}, AGIEval~\citep{zhong2023agievalhumancentricbenchmarkevaluating}, GPQA-Diamond~\citep{rein2023gpqagraduatelevelgoogleproofqa}, and RULER~\citep{hsieh2024ruler}. We evaluate RULER at the native 32K context length and, after YaRN position extrapolation, at 128K. Benchmark prompts, shot counts, decoding settings, seeds, and metric aggregation are held fixed across attention variants within each model scale.

\section{Detailed Model-Level Results}
\label{sec:appendix:model_results}

Tables~\ref{table:knowledge_benchmark}, \ref{table:reasoning_benchmark}, and \ref{table:retrieval_benchmark} report the per-task results summarized in Table~\ref{table:model_level_summary}.

\begin{table}[H]
  \centering
  \caption{
    \textbf{Knowledge Benchmark Results}.
    Knowledge evaluation of the scaling-law models at 14B and the continued-trained models at 32B. \shortName{} produces averages comparable to FullAttn at both scales and remains close across individual tasks.
  }
  \vspace{-0.75em}
  \resizebox{\linewidth}{!}{
    \begin{tabular}{@{}lccccccccccccccc@{}}
    \toprule
    \sc{Model} & \sc{MMLU} & \sc{MMLU-Pro} & \sc{BBH} & \sc{HellaSwag} & \sc{OBQA} & \sc{WinoGrande} & \sc{Avg}
    \\
    & \sc{Gen|5-shot} & \sc{5-shot} & \sc{CoT|3-shot} & \sc{5-shot} & \sc{5-shot} & \sc{5-shot} &
    \\
    \midrule
    \multicolumn{8}{c}{\text{14B-Non-Thinking, temperature=0.6, top\_k=20, top\_p=0.95, max\_gen\_toks=32768, seed=[0, 42, 233, 666, 1234]}} \\
    \midrule
    FullAttn & \underline{78.21$\pm$0.31} & \textbf{64.96$\pm$0.46} & \underline{82.59$\pm$0.48} & \textbf{80.64$\pm$0.38} & 49.4$\pm2.17$ & \underline{78.16$\pm$1.14} & \underline{72.32} \\
    MoBA & 76.34$\pm$0.46 & 58.73$\pm$0.83 & 82.3$\pm$0.41 & 80.2$\pm$0.34 & \underline{49.7$\pm$2.05} & 76.24$\pm$1.45 & 70.58 \\
    DSA & 72.72$\pm$0.53 & 59.92$\pm$0.62 & 81.62$\pm$0.63 & 80.01$\pm$0.33 & 49.53$\pm$2.24 & 77.63$\pm$1.17 & 70.23 \\
    \shortName{} (ours) & \textbf{79.04$\pm$0.28} & \underline{64.89$\pm$0.43} & \textbf{83.64$\pm$0.51} & \underline{80.3$\pm$0.31} & \textbf{49.9$\pm$2.01} & \textbf{78.48$\pm$1.23} & \textbf{72.7} \\
    \midrule
    \multicolumn{8}{c}{\text{32B-Thinking, temperature=0.6, top\_k=20, top\_p=0.95, max\_gen\_toks=32768, seed=[0, 42, 233, 666, 1234]}} \\
    \midrule
    FullAttn & \underline{77.27$\pm$0.34} & \textbf{68.57$\pm$0.41} & \underline{88.39$\pm$0.54} & \textbf{83.4$\pm$0.37} & \underline{55.0$\pm$2.23} & \underline{81.14$\pm$1.1} & \underline{75.62} \\
    \shortName{} (ours) & \textbf{77.90$\pm$0.31} & \underline{68.51$\pm$0.47} & \textbf{89.42$\pm$0.61} & \underline{83.1$\pm$0.28} & \textbf{55.8$\pm$2.22} & \textbf{81.72$\pm$0.9} & \textbf{76.07} \\
    \bottomrule
    \end{tabular}
  }
  \vspace{-1.0em}
  \label{table:knowledge_benchmark}
\end{table}

\paragraph{Knowledge performance.}
Table~\ref{table:knowledge_benchmark} shows that \shortName{} preserves performance across the evaluated knowledge benchmarks, with average scores of 72.70 versus 72.32 for FullAttn at 14B and 76.07 versus 75.62 at 32B.
The task-level pattern is consistent across the two model scales: \shortName{} scores higher on MMLU and BBH, while remaining slightly below FullAttn on MMLU-Pro and HellaSwag.
These results support retention of the evaluated knowledge capabilities under complementary long-range allocation, with modest differences across individual tasks.

\begin{table}[H]
  \centering
  \caption{
    \textbf{Reasoning Benchmark Results}.
    Reasoning evaluation of the scaling-law models at 14B and the continued-trained models at 32B. \shortName{} achieves average scores comparable to FullAttn at both scales.
  }
  \vspace{-0.75em}
  \resizebox{\linewidth}{!}{
    \begin{tabular}{@{}lccccccccccccccc@{}}
    \toprule
    \sc{Model} & \sc{PIQA} & \sc{GSM8K} & \sc{Hendrycks-Math} & \sc{ARC-C} & \sc{AGIEval} & \sc{GPQA-Diamond} & \sc{Avg}
    \\
    & \sc{5-shot} & \sc{CoT|4-shot} & \sc{CoT|4-shot} & \sc{0-shot} & \sc{0-shot} & \sc{CoT|3-shot} &
    \\
    \midrule
    \multicolumn{8}{c}{\text{14B-Non-Thinking, temperature=0.6, top\_k=20, top\_p=0.95, max\_gen\_toks=32768, seed=[0, 42, 233, 666, 1234]}} \\
    \midrule
    FullAttn & \textbf{81.63$\pm$0.86} & 90.32$\pm$0.93 & \textbf{62.46$\pm$0.65} & 61.95$\pm$1.44 & \textbf{50.56$\pm$0.71} & 39.87$\pm$3.47 & \underline{64.46} \\
    MoBA & 77.73$\pm$0.93 & \textbf{91.29$\pm$0.97} & 39.0$\pm$1.38 & \textbf{62.45$\pm$1.7} & 48.96$\pm$0.87 & 38.88$\pm$3.81 & 59.71 \\
    DSA & 80.42$\pm$0.77 & 90.52$\pm$0.99 & 55.99$\pm$0.82 & 62.04$\pm$1.29 & 47.6$\pm$0.63 & \textbf{43.11$\pm$4.52} & 63.28 \\
    \shortName{} (ours) & \underline{81.5$\pm$0.72} & \underline{90.58$\pm$0.83} & \underline{62.31$\pm$0.62} & \underline{62.1$\pm$1.62} & \underline{50.44$\pm$0.76} & \underline{42.33$\pm$3.17} & \textbf{64.87} \\
    \midrule
    \multicolumn{8}{c}{\text{32B-Thinking, temperature=0.6, top\_k=20, top\_p=0.95, max\_gen\_toks=32768, seed=[0, 42, 233, 666, 1234]}} \\
    \midrule
    FullAttn & \textbf{85.97$\pm$0.88} & \underline{93.33$\pm$0.69} & \textbf{94.94$\pm$0.82} & \underline{69.54$\pm$1.34} & \underline{54.18$\pm$0.68} & \textbf{56.06$\pm$3.54} & \textbf{75.67} \\
    \shortName{} (ours) & \underline{85.75$\pm$0.83} & \textbf{93.59$\pm$0.67} & \underline{94.71$\pm$0.84} & \textbf{69.74$\pm$1.41} & \textbf{54.36$\pm$0.65} & \underline{55.05$\pm$3.56} & \underline{75.53} \\
    \bottomrule
    \end{tabular}
  }
  \vspace{-1.0em}
  \label{table:reasoning_benchmark}
\end{table}

\paragraph{Reasoning performance.}
Table~\ref{table:reasoning_benchmark} shows that \shortName{} closely tracks FullAttn across the evaluated reasoning tasks.
Average scores are 64.87 versus 64.46 at 14B and 75.53 versus 75.67 at 32B; excluding GPQA-Diamond, all per-task differences are within 0.26 points at both scales.
The modest average gain at 14B is largely driven by GPQA-Diamond, which also exhibits relatively large run-to-run variability.
At 14B, \shortName{} additionally retains a Hendrycks-Math score close to FullAttn (62.31 versus 62.46), whereas MoBA and DSA score 39.00 and 55.99, respectively.
Overall, these results support comparable performance on the evaluated reasoning benchmarks.

\begin{table}[H]
  \centering
  \caption{
    \textbf{Retrieval Benchmark Results}.
    RULER evaluation at the native 32K context length and under YaRN extrapolation to 128K. \shortName{} achieves average scores close to FullAttn at both context lengths. NIAH-S averages NIAH-S-1, NIAH-S-2, and NIAH-S-3; NIAH-MK averages NIAH-MK-1, NIAH-MK-2, and NIAH-MK-3; and RULER-QA averages RULER-QA-SQuAD and RULER-QA-HotpotQA.
  }
  \vspace{-0.75em}
  \resizebox{\linewidth}{!}{
    \begin{tabular}{@{}lccccccccc@{}}
    \toprule
    \sc{Model} & \sc{NIAH-S} & \sc{NIAH-MK} & \sc{NIAH-MQ} & \sc{NIAH-MV} & \sc{RULER-VT} & \sc{RULER-CWE} & \sc{RULER-FWE} & \sc{RULER-QA} & \sc{Avg} \\
    \\
    \midrule
    \multicolumn{10}{c}{\text{14B Native 32K Sequence Length, seed=[0, 42, 233, 666, 1234]}} \\
    \midrule
    FullAttn & \textbf{100$\pm$0} & \textbf{98.0$\pm$0.65} & \textbf{98.8$\pm$0.68} & \underline{95.5$\pm$0.72} & \textbf{99.44$\pm$0.2} & \textbf{75.7$\pm$0.69} & 91.67$\pm$0.71 & \underline{56.28$\pm$1.2} & \textbf{89.42} \\
    MoBA & \underline{99.8$\pm$0.004} & 94.4$\pm$0.63 & 98.2$\pm$0.73 & 95.1$\pm$0.7 & 99.0$\pm$0.32 & 67.35$\pm$0.96 & 91.44$\pm$0.69 & 52.12$\pm$1.64 & 87.17 \\
    DSA & \textbf{100$\pm$0} & 95.6$\pm$0.68 & 98.0$\pm$0.77 & \textbf{96.1$\pm$0.76} & 99.24$\pm$0.26 & 69.25$\pm$0.87 & \underline{91.74$\pm$0.67} & 51.42$\pm$1.43 & 87.67 \\
    \shortName{} (ours) & \textbf{100$\pm$0} & \underline{96.96$\pm$0.71} & \underline{98.7$\pm$0.69} & 95.2$\pm$0.73 & \underline{99.28$\pm$0.21} & \underline{72.6$\pm$0.8} & \textbf{92.53$\pm$0.66} & \textbf{57.74$\pm$1.14} & \underline{89.13} \\
    \midrule
    \multicolumn{10}{c}{\text{32B Native 32K Sequence Length, seed=[0, 42, 233, 666, 1234]}} \\
    \midrule
    FullAttn & \textbf{100$\pm$0} & \textbf{99.87$\pm$0.09} & \textbf{99.8$\pm$0.08} & \textbf{99.4$\pm$0.13} & \textbf{99.64$\pm$0.18} & \textbf{85.18$\pm$0.51} & \underline{94.53$\pm$0.53} & \underline{63.2$\pm$0.93} & \textbf{92.7} \\
    \shortName{} (ours) & \textbf{100$\pm$0} & \underline{98.93$\pm$0.11} & \underline{99.69$\pm$0.09} & \underline{99.17$\pm$0.17} & \underline{99.49$\pm$0.2} & \underline{83.26$\pm$0.59} & \textbf{95.42$\pm$0.45} & \textbf{64.64$\pm$0.88} & \underline{92.58} \\
    \midrule
    \multicolumn{10}{c}{\text{14B YaRN 128K Sequence Length, seed=[0, 42, 233, 666, 1234]}} \\
    \midrule
    FullAttn & \textbf{98.4$\pm$0.51} & 55.2$\pm$1.64 & \textbf{88.25$\pm$0.69} & \underline{64.1$\pm$1.16} & \textbf{93.04$\pm$0.63} & \underline{5.76$\pm$0.46} & \textbf{84.26$\pm$0.79} & \textbf{37.73$\pm$2.15} & \underline{65.84} \\
    MoBA & 94.8$\pm$0.57 & 53.2$\pm$1.77 & 83.35$\pm$0.83 & 62.4$\pm$1.27 & 90.2$\pm$0.71 & 0.03$\pm$0.02 & 80.53$\pm$0.92 & 27.67$\pm$2.33 & 61.52 \\
    DSA & \underline{97.5$\pm$0.6} & \underline{55.4$\pm$1.62} & 84.1$\pm$0.8 & 64.0$\pm$1.14 & 92.62$\pm$0.61 & 5.42$\pm$0.86 & 80.07$\pm$1.07 & 33.34$\pm$1.87 & 64.05 \\
    \shortName{} (ours) & 97.06$\pm$0.56 & \textbf{57.6$\pm$1.44} & \underline{84.6$\pm$0.79} & \textbf{66.65$\pm$1.11} & \underline{92.88$\pm$0.63} & \textbf{13.22$\pm$1.12} & \underline{83.27$\pm$0.8} & \underline{37.54$\pm$2.04} & \textbf{66.60} \\
    \midrule
    \multicolumn{10}{c}{\text{32B YaRN 128K Sequence Length, seed=[0, 42, 233, 666, 1234]}} \\
    \midrule
    FullAttn & \textbf{99.8$\pm$0.12} & \textbf{86.4$\pm$1.35} & \textbf{95.7$\pm$0.56} & \underline{88.0$\pm$0.79} & \textbf{98.76$\pm$0.22} & \underline{45.68$\pm$1.03} & \textbf{93.27$\pm$1.25} & \textbf{48.61$\pm$2.17} & \textbf{82.03} \\
    \shortName{} (ours) & \underline{99.6$\pm$0.15} & \underline{84.3$\pm$1.47} & \underline{95.1$\pm$0.6} & \textbf{88.5$\pm$0.81} & \underline{98.53$\pm$0.25} & \textbf{48.23$\pm$0.94} & \underline{91.84$\pm$1.31} & \underline{48.11$\pm$2.13} & \underline{81.78} \\
    \bottomrule
    \end{tabular}
  }
  \vspace{-1.0em}
  \label{table:retrieval_benchmark}
\end{table}

\paragraph{Long-context performance.}
Table~\ref{table:retrieval_benchmark} shows that \shortName{} retains aggregate retrieval performance comparable to FullAttn at both the native 32K context length and under YaRN extrapolation to 128K.
At 128K, its average score is 66.60 versus 65.84 for FullAttn at 14B, and 81.78 versus 82.03 at 32B.
Individual tasks exhibit both gains and losses.
Some differences are small relative to the reported run-to-run evaluation variability, whereas others are more pronounced, such as the lower NIAH-MQ score and higher RULER-CWE score at 14B.
These results support comparable aggregate long-context performance, while indicating task-specific trade-offs rather than uniform preservation of every retrieval capability.

\end{document}